\documentclass{article}

    \PassOptionsToPackage{numbers, compress}{natbib}

\usepackage[preprint]{neurips_2024}

\usepackage[utf8]{inputenc} 
\usepackage[T1]{fontenc}    
\usepackage{hyperref}       
\usepackage{url}            
\usepackage{booktabs}       
\usepackage{amsfonts}       
\usepackage{nicefrac}       
\usepackage{microtype}      
\usepackage{xcolor}         
\usepackage{enumitem}

\usepackage[utf8]{inputenc} 
\usepackage[T1]{fontenc}    
\usepackage{hyperref}       
\usepackage{url}            
\usepackage{booktabs}       
\usepackage{amsfonts}       
\usepackage{nicefrac}       
\usepackage{microtype}      
\usepackage{xcolor}         
\usepackage{graphicx} 
\usepackage{enumitem} 
\usepackage{amsmath} 

\usepackage[utf8]{inputenc}
\usepackage{graphicx}
\usepackage{booktabs}
\usepackage{xcolor}   
\usepackage{array}
\definecolor{Gray}{gray}{0.9}

\usepackage[utf8]{inputenc}

\usepackage{amsmath}
\usepackage{amssymb}

\usepackage{xcolor}

\usepackage{wrapfig}

\title{Curriculum Learning with Quality-Driven Data Selection }

\author{%
  Biao Wu \\
  Australian Artificial Intelligence Institute \\ 
  \texttt{biaowu165534@gmail.com} \\
  \And
  Ling Chen \\
  Australian Artificial Intelligence Institute \\ 
  \texttt{Ling.Chen@uts.edu.au } \\
}

\begin{document}

\maketitle

\begin{abstract}

The remarkable multimodal capabilities demonstrated by OpenAI’s GPT-4 have sparked significant interest in the development of Multimodal Large Language Models (MLLMs). Visual instruction tuning of MLLMs using machine-generated instruction-following data has been shown to improve zero-shot capabilities on many tasks, but there has been less exploration of controlling the instruction data quality. Current methodologies for data selection in MLLMs often rely on single, unreliable scores or use downstream tasks for selection, which is time-consuming and can lead to potential overfitting on the chosen evaluation datasets. To mitigate these limitations, we propose a novel data selection methodology that utilizes image-text correlation and model perplexity to evaluate and select data of varying quality. This approach leverages the distinct distribution of these two attributes, mapping data quality into a two-dimensional space that allows for the selection of data based on their location within this distribution. By utilizing this space, we can analyze the impact of task type settings, used as prompts, on data quality. Additionally, this space can be used to construct multi-stage subsets of varying quality to facilitate curriculum learning. This multiple training strategy not only utilizes a minimal amount of data but also maintains data quality diversity, significantly enhancing the model’s fine-tuning performance. Our research includes comprehensive experiments conducted on various datasets. The results emphasize substantial enhancements in five commonly assessed capabilities compared to using the complete dataset. Our codes, data, and models are publicly available at: \url{https://anonymous.4open.science/r/EHIT-31B4}

\end{abstract}


%

\section{Introduction}

Instruction-following Multimodal Large Language Models (MLLMs) excel in multi-modality tasks \cite{llava1.5,zhao2023svit,nguyen2024improving}. Their effectiveness largely comes from using Large Language Models (LLMs) to generate synthetic data for visual instruction tuning. SELF-FILTER \cite{chen2024your} emphasizes that visual instruction tuning is a straightforward alignment process in MLLMs training. It only needs a small amount of tuning data to activate the pre-trained capabilities and align them with the target interaction format. To improve this process, dataset selection tasks have been proposed to choose high-quality instruction-tuning data, enhancing the performance of these models \cite{chen2024your}. 



Despite the central role that datasets play in training large language models, exploring data quality for instruction tuning in vision-and-language models remains challenging. Many existing data selection methods use simple rules based on the characteristics of images and texts separately, such as the length of captions, the use of nouns, the complexity of sentences, the aspect ratio of images, and the minimum size of images \cite{kakaobrain2022coyo-700m, schuhmann2022laion, changpinyo2021conceptual, sharma2018conceptual}. These methods also consider the reliability of the data source \cite{desai2021redcaps}. More advanced techniques focus on the alignment between images and texts, using models like CLIP \cite{openclip} to evaluate how closely the content of an image matches the accompanying text. This is done by measuring the similarity between image and text features \cite{radford2021learning,schuhmann2022laion,nguyen2024improving} or by checking if the image's main object is mentioned in the caption \cite{sharma2018conceptual}.  However, These approaches focus on high-quality data, with limited exploration of data quality diversity.

To improve the effectiveness of multimodal instruction data selection and the utilization of data diversity, we propose a new data selection method. This method constructs a representation space through two attributes of the data, which allows clear observation of the data in distributional differences for different task type settings. Meanwhile, this effectively categorizes data quality by distribution, allowing us to select different quality subsets for training. Specifically, our new method calculates each sample's clip score and model loss, using them as two-dimensional coordinates. By dividing key areas, we can obtain subsets of data with varying quality.

We introduce a new training strategy, curriculum learning. Unlike most instruction tuning tasks, our curriculum learning method involves multiple training stages, each using progressively higher-quality data. We begin by training on data randomly sampled from a high-quality sample space. In subsequent stages, we progressively refine the distribution area of this high-quality data, sampling from increasingly focused spaces. This iterative training process mimics the human learning approach, enabling the model to use data of varying quality to maintain diversity. Through extensive experiments on LLaVA-v1.5, we demonstrate that our methods can surpass models trained on the full instruction data using only about 5\% of the raw instruction tuning dataset samples. This improvement is consistent across multiple evaluation datasets and benchmarks.

We summarize the main contributions of this paper:

\begin{enumerate}[leftmargin=*]  
    \item We propose a new method for selecting high-quality data by observing its quality distribution and creating a subset. We found a correlation between data quality and its distribution. Additionally, the task type for text generation impacts the quality of visual instruction tuning data.
    
    
    
    \item We propose a curriculum learning strategy and demonstrate that the model’s performance can be further improved by using a multi-stage approach to adjust the quality of the training data, requiring only a small amount of data.
   
    \item We evaluated our method on multiple tasks and achieved better performance using only 5\% compared to having used the full data.
\end{enumerate}



\vspace{-2mm}
\section{Related Work}
\vspace{-2mm}

\paragraph{Multimodal Instruction Tuning} Multimodal Instruction Tuning is pivotal in advancing the capabilities of models like LLaVA \cite{LLaVA}, MiniGPT-4  \cite{MiniGPT4}, and InstructBLIP \cite{instructblip}, which thrive on intricately paired image-text data. This technique refines the models' performance beyond what is achievable with conventional VQA datasets \cite{goyal2017making, GQA2019}, which often provide limited, short-answer data that can impair model performance. Recognizing this, the MiniGPT-4 \cite{MiniGPT4} team curated a dataset of 3,500 image-text pairs, refined through interactions with ChatGPT, to enhance the models' ability to generate nuanced responses. Similarly, LLaVA \cite{LLaVA} set a benchmark by creating LLaVA-Instruct-150K, a dataset generated by prompting GPT-4 with rich annotations from the COCO dataset \cite{MSCOCO}, including image captions and object details, to produce detailed questions and answers. Expanding the scope, LLaVAR \cite{LLaVAR} addressed the challenges of interpreting text-rich images by assembling over 422,000 pieces of instruction-following data through OCR technology, supplemented by an additional 16,000 high-quality entries processed by GPT-4. Furthermore, InstructBLIP \cite{instructblip} incorporated a diverse array of 26 public datasets, including LLaVA-Instruct-150K, to create a more comprehensive visual instruction tuning dataset. This effort, however, highlighted the prevalence of brief, perceptually focused content in existing datasets. Meanwhile, M$^3$IT \cite{M3IT}  transformed 40 distinct datasets into a unified vision-to-text framework, utilizing ChatGPT to rephrase and enrich the context of the responses, broadening the scope of training data suitable for deep learning models. This collective endeavor to enrich multimodal datasets \cite{VisualGenome,li2023stablellava}  illustrates a strategic pivot towards generating a larger, more varied corpus of visual instruction data. These datasets now cover an extensive range of tasks from basic visual recognition to complex reasoning and planning, setting a new standard for training sophisticated multimodal systems.

\paragraph{Data Selection} Data selection is a developing field in the instruction-tuning of large language models, focused on identifying high-quality data and removing harmful information that could lead to errors ~\cite{chen2023alpagasus, cao2023instruction}. In this area,  ~\cite{chen2023alpagasus} introduced Alpagasus, a method that automates data selection by assessing instruction quality via queries to ChatGPT, thereby improving training efficiency. \cite{li2023quantity}  suggested using the IFD score as an indicator of data difficulty, while ~\cite{cao2023instruction} developed Instruction Mining, which evaluates sample quality through a linear combination of various indicators. Concurrently, \cite{li2023one} proposed assessing data by the one-shot learning performance on specific tasks. Finally, ~\cite{wei2023instructiongpt} in their study on InstructionGPT-4, apply a combination of multimodal scores and a regression model trained on predefined tasks for data selection, although their application is confined to MiniGPT-4 \cite{MiniGPT4}, which includes just 3,400 instructions.

\paragraph{Curriculum Learning} 
Curriculum Learning has emerged as an effective strategy in machine learning, allowing models to start with simpler tasks and gradually progress to more complex ones. This method, inspired by the way humans learn, has been applied across various domains such as natural language processing and computer vision ~\cite{bengio2009curriculum,soviany2022curriculum}. In this context, ~\cite{bengio2009curriculum} pioneered the concept by showing how a progressive learning schedule can improve performance in neural networks. More recently, ~\cite{soviany2022curriculum} proposed a dynamic curriculum learning approach that adjusts the difficulty of the data based on the model's performance during training. Additionally, ~\cite{ma2022curriculum} introduced an automatic curriculum learning framework that utilizes reinforcement learning to dynamically select training samples, optimizing the learning process. Lastly, ~\cite{soviany2021curriculum} explored self-paced learning, a variation where the model self-assesses and chooses the appropriate learning pace, thereby aligning with curriculum learning principles to improve overall training efficacy.


\section{Methods}

\subsection{Data selection}

We define our data selection task in the context of instruction fine-tuning. Given an instruction tuning dataset \(\mathcal{D} = \{\mathbf{x}_j\}_{j=1}^{N}\), where each \(\mathbf{x}_j = (\mathbf{x}_j^i, \mathbf{x}_j^t)\) represents a pair of input image and text, our objective is to select a subset of size \(m\) from \(\mathcal{D}\). The goal is to prune \(\mathcal{D}\) such that the resulting subset, \(\mathcal{D}_{f}^{m} \subset \mathcal{D}\), enables the pre-trained vision-language model \(f\) to achieve optimal performance on downstream tasks \(\{T_i\}_{i=1}^t\). Here, \(|\mathcal{D}_{f}^{m}| = m\). 


\subsection{Data Curriculum}

We propose to select a subset of the dataset based on 1) clip score for image-text feature similarity and 2) model loss for data perplexity. We use these two data attributes to create a representation space for all data instances. By employing this method, we select the data using the region of the representation space that exhibits higher or lower values for both attributes. The vision-language model \(f\) is pre-trained.  We denote its total loss as \(l\) and the loss on visual instruction data \(x_i\) as \(l_i\). Additionally, We denote its clip score as \(s\) and the correlation on visual instruction data \(x_i\) as \(s_i\).  By maximizing the  \(l\) and  \(s\), we can obtain a relatively high-quality subset of data \(\mathcal{D}_{f}^{m}\).

\paragraph{Intermediate Data Similarity}

To evaluate the similarity between an input image \(x_j^i\) and text \(x_j^t\), we use the CLIP model to extract features from both. Specifically, we apply the image encoder of the CLIP, defined as \( I ( \cdot ) \), to obtain the feature vector from the image, and the text encoder of the CLIP, defined as \( T ( \cdot ) \), to derive the feature vector from the text. We then compute the dot product of both features to generate a clip score, which we define as \( s_j \).

\vspace{-4mm}
\[
s_j = I ( x_j^i ) \cdot   T ( x_j^t )
\]

We partition the data subset by identifying the upper bounds \(S_{max}\) and lower bounds \(S_{min}\) of the \( s_j \). Using these bounds, we select the sample data \(d_j\) to obtain the corresponding subset, which we refer to as the Data of Intermediate Similarity (DIS):

\vspace{-3mm}
\[
DIS =  \{d_j \mid S_{min} \leq s_j \leq S_{max}\}
\]

Clip score reflects how well the image features correspond to the text features, allowing us to identify and select high-quality data where the image and text are closely related. 

\paragraph{Intermediate Data Loss} 

The loss produced by the model, which is also a measure of perplexity, reflects the difference between the target text and the model's internal preferences. A higher loss makes the learning process more challenging for the model. Following a standard LLaVA architecture, the image encoder provides latent encoded features \(X_j\). Concurrently, the text decoder is tasked with maximizing the conditional likelihood of the paired text \(Y_j\) under the forward autoregressive factorization:

\vspace{-4mm}
\[
l_j = -\sum_{t=1}^T \log P_{\theta}(Y_{j,t} \mid Y_{j,<t}, X_j)
\]
\vspace{-3mm}


We partition the data subset by detecting the upper bounds \(L_{max}\) and lower bounds \(L_{min}\) of the loss. Using these bounds, we select the sample data \(d_j\) to obtain the corresponding subset, which we refer to as the Data of Intermediate Loss (DIL):

\vspace{-3mm}
\[
DIL =  \{d_j \mid  L_{min} \leq l_j \leq L_{max}\}
\]

\paragraph{Intermediate Data  Quantity} When each piece of data has clip score and loss, we can construct a two-dimensional representation space based on these two attributes. Therefore, we select the sample data \(d_j\) and set both related upper bounds and lower bounds to select the high-quality subset, which we refer to as the Data of Intermediate Quantity (DIQ) :

\vspace{-3mm}

\[
 DIQ = \{d_j \mid  L_{min} \leq l_j \leq L_{max}\  ,  S_{min} \leq s_j \leq S_{max}\}
\]


We propose a data curriculum framework that starts training with simpler tasks and progressively advances to more complex ones. Based on our 
$ DIQ $, we divide the region into unified blocks and use $\Delta L$ and $\Delta S$ , corresponding to model loss and clip score respectively. By employing data selection methods, we can control the quality of a subset of data by gradually increasing clip score thresholds and loss thresholds. Consequently, we divide the learning process into several phases $k$, and we select the sample data in each phase with different quantities :

\[
 C_k = \{d_j \mid  L_{p} \leq l_j   ,  S_{p} \leq s_j  \}
\]
where
\[
   L_{p} = L_{min} + k \Delta L ~\text{and}~ S_{p} = S_{min} + k \Delta S.
\]

As k increases, the learning process can be divided into multiple phases: Initialization, Intermediate, and Advanced.

\begin{itemize}[leftmargin=*]  
  \item Initialization Phase (k=0): The model starts with a distribution of high-quality data, focusing on underlying patterns without being overwhelmed by complexity.
  \item Intermediate Phase (k=1): Data quality is improved by increasing the thresholds for clip score and loss, narrowing the candidate region of high-quality data.
  \item Advanced Phase (k=2): The model is exposed to the most challenging data, characterized by higher clip score and model loss, testing its ability to handle complex and less consistent relationships.
\end{itemize}

This phased approach ensures progressive learning, better generalization, reduced overfitting, and enhanced robustness. By systematically organizing and presenting data based on quality metrics, the data curriculum ensures the model develops a solid foundation before tackling more complex data, leading to improved performance on multimodal tasks.

\section{Experiments} 

In this section, we first detail our settings and the chosen base models. Then we introduce the different train scenarios and evaluation benchmarks used in our experiments and the baseline methods.  We show that our proposed method achieves better performance on multiple tasks using less data.

\subsection{Experimental Setup}


\paragraph{VL instruction data.}
We use the core set, SVIT-core-157K, as our raw data, totaling 157,712 samples. SVIT \cite{zhao2023svit} extends visual instruction tuning data to present a large-scale dataset containing 4.2 million command adjustment data. These data include dialog Q\&A pairs, complex inference Q\&A pairs, referring Q\&A pairs, and detailed descriptions.  More details can be found in the Appendix.

\paragraph{Base models.} We use the LLaVA-v1.5-7B \cite{llava1.5} model architecture and its pre-training weights as our base models. The entire LLaVA training process is divided into two stages. For the first stage of pretraining, LLaVA-1.5-558k \cite{LLaVA} selected from CC3M data are used, which have been converted into instruction-following data by GPT-4. For the second stage of visual instruction tuning, LLaVA-1.5-mix-665k \cite{llava1.5} has been used.  

\paragraph{Train setting}
We consider LoRA finetuning for the new instruction data. We define the state where LLaVA-1.5-mix-665k \cite{llava1.5} has been used for instruction tuning as scenario 1, and the state where this data has not yet been used for instruction tuning as scenario 2.  And, to verify the effectiveness of the data selection strategy for LLaVA model training, we mainly consider these two scenarios.

\paragraph{Benchmarks}
We assess our methods using a mix of academic-task-oriented benchmarks and new benchmarks tailored for instruction-following LMMs, covering a total of 5 benchmarks. For academic-focused benchmarks, VQA-v2~\cite{goyal2017making} and GQA~\cite{GQA2019} test the model's visual perception abilities with open-ended questions. VizWiz~\cite{gurari2018vizwiz} includes 8,000 images to evaluate the model's zero-shot generalization on visual queries from visually impaired individuals. In line with InstructBLIP~\cite{dai2023instructblip}, we use the image subset of ScienceQA~\cite{lu2022learn} with multiple-choice questions to gauge zero-shot performance in scientific question answering. TextVQA~\cite{singh2019towards} involves text-rich visual question answering.

\subsection{Scenario 1: Training from LLaVA} 

We use the LLaVA-v1.5-7B \cite{llava1.5} architecture with model weights fully fine-tuned using LLaVA-1.5-mix-665k data. Subsequently, we fine-tune this model with LoRA \cite{lora} during the follow-up experiments. In training, we keep the visual encoder, projector, and LLM weights frozen, and maximize the likelihood of with trainable parameters of LoRA only.  We keep the rest of the training protocol the same to allow for a fair comparison. Scenario 1, which only includes LoRA tuning, takes approximately 16 hours on an NVIDIA Tesla A100 GPU with 40GB of memory, using DeepSpeed ZeRO Stage 3.
We use the SVIT-core-157K \cite{zhao2023svit} dataset for continuous fine-tuning to establish a baseline. And the same method is applied to fine-tune our data.

\begin{table*}[ht]
\centering
\scalebox{0.9}{
\begin{tabular}{l l p{5mm}p{7mm}p{9mm} | p{8mm}p{8mm}p{8mm}p{8mm}p{8mm}}
\toprule
Method & LLM & Res. & PT & IT & VQA$^\text{v2}$ & GQA & VisWiz & SQA$^\text{I}$ & VQA$^\text{T}$ \\
\midrule
BLIP-2\cite{Blip-2} & Vicuna-13B & 224 & 129M & - & 41.0 & 41 & 19.6 & 61 & 42.5 \\
InstructBLIP\cite{instructblip} & Vicuna-7B & 224 & 129M & 1.2M & -- & 49.2 & 34.5 & 60.5 & 50.1 \\
InstructBLIP\cite{instructblip} & Vicuna-13B & 224 & 129M & 1.2M & -- & 49.5 & 33.4 & 63.1 & 50.7 \\
Shikra\cite{shikra} & Vicuna-13B & 224 & 600K & 5.5M & 77.4 & -- & -- & -- & -- \\
IDEFICS-9B \cite{idefics} & LLaMA-7B & 224 & 353M & 1M & 50.9 & 38.4 & 35.5 & -- & 25.9 \\
IDEFICS-80B\cite{idefics} & LLaMA-65B & 224 & 353M & 1M & 60.0 & 45.2 & 36.0 & -- & 30.9 \\
Qwen-VL\cite{bai2023qwen} & Qwen-7B & 448 & 1.4B$^\dagger$ & 50M$^\dagger$ & 78.8 & 59.3 & 35.2 & 67.1 & 63.8 \\
Qwen-VL-Chat\cite{bai2023qwen} & Qwen-7B & 448 & 1.4B$^\dagger$ & 50M$^\dagger$ & 78.2 & 57.5 & 38.9 & 68.2 & 61.5 \\
LLAVA-V1.5\cite{llava1.5} & Vicuna-7B & 336 & 558K & 665K & 78.5 & 62.0 & 50.0 & 66.8 & 58.2 \\
\midrule
+ SVIT-Core-157K\cite{zhao2023svit} & Vicuna-7B  & 336 & 558K & +157K & {75.9} & {57.1} & {49.1} & 69.0 & 56.3 \\
+ Ours  & Vicuna-7B  & 336 & 558K & +7K & {77.9}  & {61.8} & {51.1} & 69.5 & 57.3  \\ 
\bottomrule
\end{tabular}
}
\caption{\textbf{Comparison with SoTA methods on 5 benchmarks.} We achieves better performance on all benchmarks than SVIT-Core-157K. Res, PT, and IT indicate input image resolution, and the number of samples in the pretraining and instruction tuning stage, respectively.
Benchmark names are abbreviated due to space limits.  VQA-v2 \cite{goyal2017making}, GQA \cite{GQA2019}, VisWiz \cite{gurari2018vizwiz}, ScienceQA-IMG \cite{lu2022learn}, TextVQA \cite{singh2019towards}. More details can be found in the Evaluation Metrics section of the Appendix.}
\label{tab:results}
\end{table*}



We report our main results in Table \ref{tab:results}. Our method, using only 7000 samples of  SVIT-core-157K, achieved higher performance across all benchmarks compared to the full data experiment setup. Furthermore, it surpassed the base model on SQA \cite{lu2022learn} and VisWiz \cite{gurari2018vizwiz}, reaching state-of-the-art (SOTA) performance. In the efficient LoRA training setup, our data exceeded SVIT-core-157K\cite{zhao2023svit} by 4.7 points in GQA \cite{GQA2019}, 2.0 points in VQAV2 \cite{goyal2017making}, 1.0 point in TextVQA \cite{singh2019towards}, 2.0 points in VisWiz \cite{gurari2018vizwiz}, and 0.5 points in SQA \cite{lu2022learn}. The improvements verify the better training effects of our data since less data amount and same model are used.


\paragraph{Effectiveness of DIQ} 

In Table 2, we use the top-right corner in the left panel of Figure 7 (shown in the appendix) as the top 5\% of the DIQ and conducted a comparison experiment, we found that using the 5\% selected by DIQ resulted in better performance compared to using the top 5\% of DIS and DIL separately.  

\begin{wraptable}{r}{0.45\textwidth} 
\centering

\vspace{1mm}
\label{tab:results_scenario_1}
\small
\scalebox{1}{
\begin{tabular}{c|ccc}
\toprule
\textbf{Strategy} & \multicolumn{3}{c}{\textbf{Scenario 1}} \\
                 & SQA   & TextVQA & GQA   \\  
\midrule  
DIS              & 57.06 & 56.13  & 61.06 \\
DIL              & 68.82 & 56.30  & 60.87  \\ 
DIQ              & \textbf{69.56} & 56.84  & 61.16  \\ 
\midrule 
\multicolumn{4}{l}{\it Result with Data Curriculum } \\  
Ours             & 69.51 & \textbf{57.25} & \textbf{61.80}  \\ 
\bottomrule
\end{tabular}
}

\caption{Results across different methods.}
\end{wraptable}

We realized that this improvement is due to the subset from DIQ selecting data evenly from the entire region, whereas DIS and DIL focus on regions with high levels of clip score or loss. Based on these insights, we introduced curriculum learning, utilizing multi-stage training that progresses from low-quality to high-quality data. This approach, as demonstrated in the ablation experiment in Table 2, highlights the importance of increasing the diversity of data quality for improving model performance. By employing this method,  we found that using curriculum learning with the DIQ method can further enhance model performance.

To further understand the effectiveness of curriculum learning, we observe that it starts with simple examples, which have lower noise and smaller loss. This provides a smoother loss landscape, reducing gradient oscillations and instability for a more stable initial training process. As the model progresses to higher-quality data, it benefits from established initial parameters and a clear learning direction, facilitating easier optimization. By gradually increasing data quality, curriculum learning helps the model adapt and optimize progressively, leading to improved performance as shown in our results.

\subsection{Scenario 2: Training from Vicuna + projection.}

To check the quality of our selected data and ensure consistency in our experiments, we use the LLaVA-v1.5-7B \cite{llava1.5} model architecture and its pre-training weights, only a projector.  We utilize this projector, the pre-trained CLIP visual encoder ViT-L/14, and Vicuna-7b to establish the weights of LLAVA that only the alignment task has been completed. This setup helps us observe how different selected datasets activate the model's ability to engage in dialogue while avoiding interference from other instruction-tuning data on this task. The rest of the model training protocol is kept unchanged for fair comparison. We keep the training setting as same as scenario 1 and only update the LoRA weights of the LLM. 

\begin{figure}[h]
\raggedright
\includegraphics[width=1\textwidth]{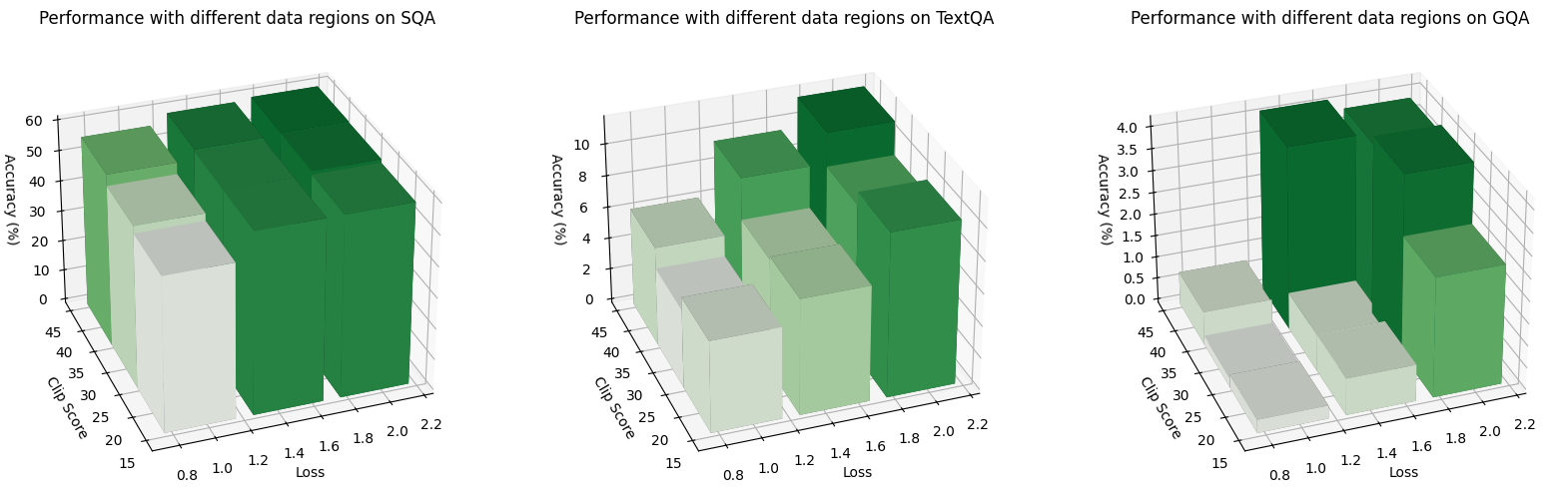} 
\caption{Comparison of ablation results with data from different DIQ regions in scenario 2. }
\label{fig:exampleImage_fig4}
\end{figure}

\paragraph{Effectiveness of DIQ} 

To verify the effectiveness of DIQ, we analyze the clip score and loss for all the data. 
In Figure 7, we divided the data into 9 regions and selected 7,000 samples from each region as corresponding data subsets using the DIQ method. The axes in Figure \ref{fig:exampleImage_fig4}  are the loss value and the clip score value. It shows the position of the columns in each region shows the range of the corresponding dual attributes. The color of the columns also reflects the size of the corresponding value, the higher the performance the darker the color of the columns, and vice versa. Combining the performance of SQA, TextVQA, and GQA, we find that data with a higher clip score and loss show better performance on the downstream task, implying that the top-right DIQ subset contains higher quality data.
%

\subsection{Exploring Different Data Selection} 

\paragraph{Effectiveness of DIS and DIL.}

\begin{figure}[h]
\raggedright
\includegraphics[width=1\textwidth]{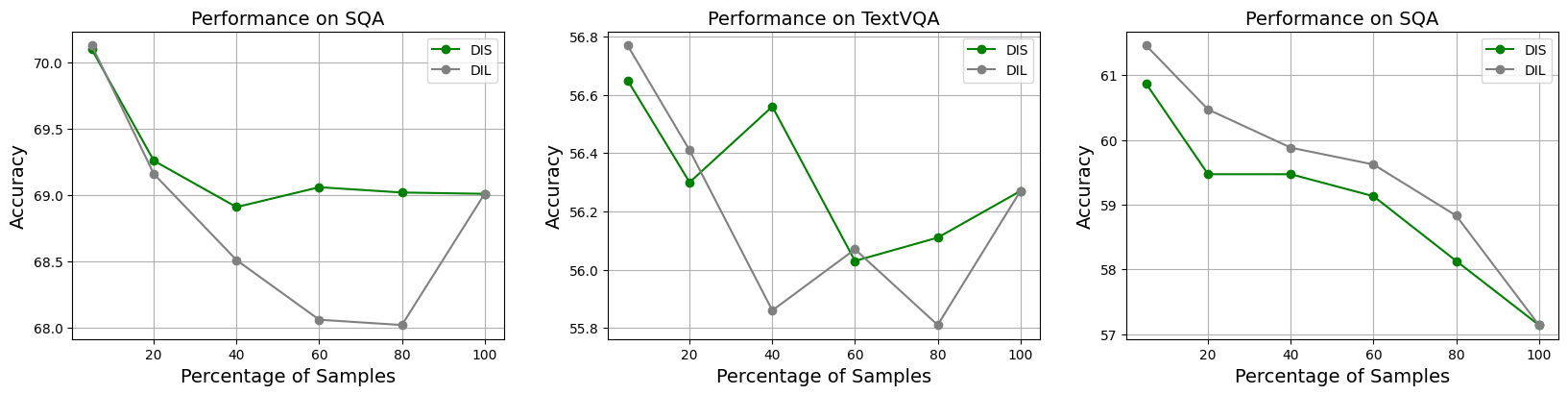} 
\caption{Comparison of ablation experiment results in scenario 1 with different data select ratios.  }
\label{fig:exampleImage_fig5}
\end{figure}

To verify the effectiveness of DIS and DIL separately, we first verified the data selection results of individual methods, in scenario 1 of the LLaVA training program. As shown in Figure \ref{fig:exampleImage_fig5}, both DIS and DIL, using only the top 5\% (around 7000 samples) of the selected data, significantly outperform the results using all the data. The model performance gradually decreases as the amount of data increases. 

\begin{figure}[h]
\raggedright
\includegraphics[width=1\textwidth]{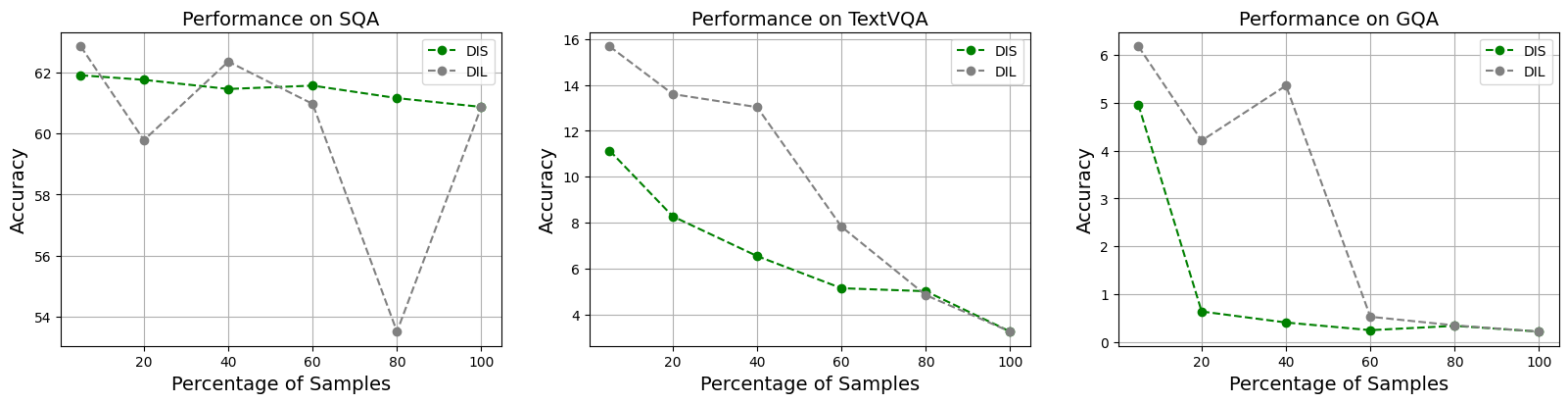} 
\caption{Comparison of ablation experiment results in scenario 2 with different data select ratios.}
\label{fig:exampleImage_fig3}
\end{figure}

As shown in Figure \ref{fig:exampleImage_fig3}, similar to scenario 1,  both DIS and DIL, using only the top 5\% of the selected data, significantly outperform the results using all the data. This result is consistent with the hypothesis presented in LIMA's \cite{zhou2024lima} study, which demonstrates that alignment also be a straightforward process in MLLM training. In this process, the model learns the style or format of interacting with users, effectively utilizing the knowledge and capabilities it acquired during pretraining. Meanwhile, a high-quality subset of data is sufficiently informative to help the model adapt well to new user interaction styles in scenario 2, compared to the full data.

\paragraph{Effectiveness of Mixed Methods.} 

Table \ref{tab:results_model_2} first compares the performance differences of the top 5\% of DIS, DIL, and DIQ in scenarios 1 and 2. We notice that using the 5\% selected by DIS and DIL separately outperformed the top 5\% of DIQ in scenario 2. We realized that this improvement is due to the DIS and DIL subsets focusing on regions with a higher clip score or loss, where data with both high attributes predominate, resulting in an overall higher data quality compared to DIQ. Based on these insights, we explore the mixed method, We combined the top high-quality subsets obtained from different methods to create a larger, high-quality subset.

Therefore, we observed that for Scenario 1, the model performs best with only 5\% of data based on DIQ. Comparing different data subsets from various regions, as well as combining data from different regions, did not improve the model's performance. That indicates that scenario 1 mainly benefits from smaller, high-quality data. For Scenario 2, the model performs best with 15\% of data based on the mix of different region data. In our comparison, we found that when increasing the data size from 5\% to 10\% with a single strategy, the performance of both DIS and DIL decreased due to a relative drop in data quality. However, when multiple top 5\% data subsets were combined, the model's performance improved, even at the same 10\% scale. This demonstrates that in scenario 2, the model relies more on the quantity of high-quality data. Consequently, when we combined the top 5\% subsets from all three regions, the model's performance improved further, confirming this observation.

\paragraph{Effectiveness of Curriculum Learning.}

In Table \ref{tab:results_model_2}, first, we randomly sampled 2,400 examples from all the regions from DIQ for the first training phase, corresponding to $ C_1 $ . In the second phase, we narrowed the range of high-quality data and randomly sampled 2,400 examples to further fine-tune the model trained in the first phase, corresponding to $ C_2 $. We repeated this process for the third phase and got $ C_3 $. In total, 7000 samples of data were used, which is consistent with the data size of the DIQ approach.  
After introducing curriculum learning in scenario 1, the model's performance improved further. However, even with curriculum learning, the model’s performance declined as the data size increased. This indicates that in scenario 1, in addition to enhancing data quality diversity, it is also crucial to maintain a small scale. For scenario 2, the model's performance further improved when using 15\% of the data. This proves that both curriculum learning and the quantity of high-quality data are important to scenario 2.

\begin{table}[h]
\centering
\scalebox{0.8}{
\begin{tabular}{c|c|cccc|cccc}
\toprule
\textbf{Strategy} & \textbf{Data Size} & \multicolumn{4}{c|}{\textbf{Scenario 1}} & \multicolumn{4}{c}{\textbf{Scenario 2}} \\
                 &                    & \textbf{SQA}   & \textbf{TextVQA} & \textbf{GQA}  & \textbf{AVG}  & \textbf{SQA}   & \textbf{TextVQA} & \textbf{GQA}  & \textbf{AVG}  \\  
\midrule
\multicolumn{10}{l}{\textit{Result with scaling the high-quality data with different subset}} \\  
5\% in DIS               & 7 k                & 57.06 & 56.13  & 61.06 & 58.08 & 61.92 & 11.14  & 4.96 & 25.34 \\
5\% in DIL               & 7 k                 & 68.82 & 56.30  & 60.87 & 62.00 & 59.79 & \underline{\textbf{15.68}}  & 6.18 & 27.22 \\
10\% in DIS               & 14 k               & 69.31 & 56.30  & 59.93 & 61.18 & 61.18 & 8.88   & 1.03 & 23.03 \\
10\% in DIL               & 14 k              & 69.46 & 55.98  & 60.42 & 61.95 & 61.63 & 13.07  & 4.76 & 26.49 \\
5\% in DIQ               & 7 k                & 69.56 & \underline{56.84}  & \underline{61.16} & \underline{62.52} & 61.53 & 11.45  & 3.89 & 25.62 \\
5\% DIS + 5\% DIL         & 14k          & 69.76 & 56.41  & 60.04 & 62.07 & 61.78 & 13.68  & 5.86 & 27.11 \\
5\% DIS + 5\% DIQ         & 14k           & 69.96 & 56.32  & 60.72 & 62.33 & 61.87 & 13.19  & 7.16 & 27.41 \\
5\% DIL + 5\% DIQ         & 14k          & 69.06 & 56.05  & 60.75 & 61.95 & 60.98 & 14.71  & 5.95 & 27.21 \\
5\% DIS + 5\% DIL + 5\% DIQ   & 21k     & \underline{\textbf{70.25}} & 56.32 & 60.72 & 62.43 & \underline{61.92} & 13.17  & \underline{\textbf{7.20}} & \underline{27.43} \\
\midrule
\multicolumn{10}{l}{\textit{Result with Curriculum Learning with different size}} \\  
C$_1$ + C$_2$ + C$_3$ (Ours)  & 7 k                & 69.51 & \textbf{57.25} & \textbf{61.80} & \textbf{62.85} & 59.93 & 9.18   & 1.78 & 23.63 \\
C$_1$ + C$_2$ + C$_3$    & 14 k               & 69.56 & 56.99  & 61.73 & 62.76 & 61.53 & 12.93  & 3.70 & 26.05 \\
C$_1$ + C$_2$ + C$_3$    & 21 k               & 69.51 & 56.54  & 61.38 & 62.48 & \textbf{61.97} & 15.49 & 6.39 & \textbf{27.95} \\
\bottomrule
\end{tabular}
}
\vspace{3mm}
\caption{Comparison of ablation results with different data selection strategies and curriculum sizes. The underlined data is the maximum value considering only scaled high-quality data, and the bolded data is the global maximum value. All the "x\%" refers to selecting top x\% examples from the corresponding data.}
\label{tab:results_model_2}
\end{table}

\vspace{-3mm}

\subsection{What Makes Selected Data Quality Different?}



Visual instruction data generated via unimodal LLM exhibit different properties on the epistemic evidence space of clip score and loss by forming image-text pairs with their corresponding images. We try to understand what causes this problem with visual instruction data and how to control the distribution and quality of visual instruction data. The existing methods for generating data for visual instruction-tuning primarily use single-mode LLMs to adjust the text format in the data. This approach can lead to inconsistencies between the images and the corresponding text content, causing mismatches or failing to accurately capture the main elements of the images. Additionally, the design of prompts often influences the visual instruction data, altering the generation process to suit different tasks. This variation in text generation methods for different tasks exacerbates the issue of data quality divergence. To better compare the distribution of data for different tasks in space, we visualize the space.

\begin{figure}[h]
	\begin{minipage}{0.48\linewidth}
		\vspace{3pt}
		\centerline{\includegraphics[width=\textwidth]{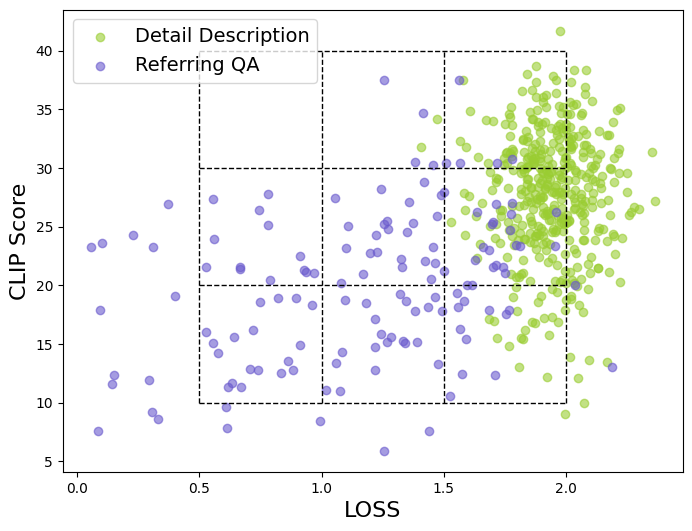}}
	    \caption{Data distribution comparison for Referring QA and Detail Description tasks.}
            \label{fig:results_human2}
	\end{minipage}
        \hspace{3mm}
	\begin{minipage}{0.48\linewidth}
		\vspace{3pt}
		\centerline{\includegraphics[width=\textwidth]{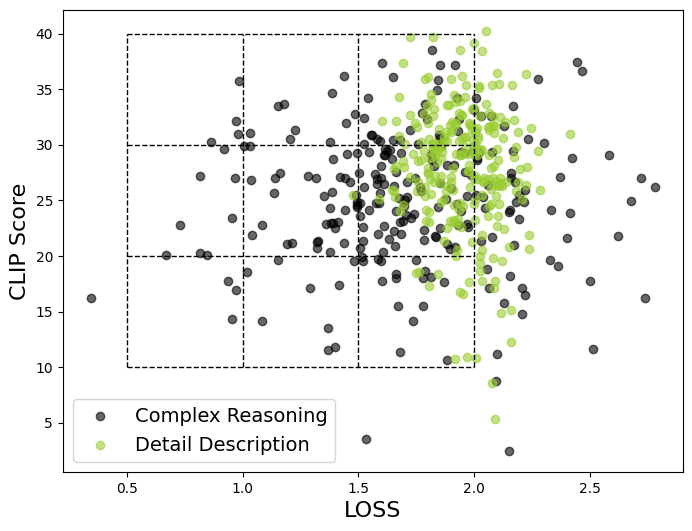}}
	      \caption{Data distribution comparison for Complex Reasoning and Detail Description tasks.}
            \label{fig:results_human3}
	\end{minipage}
	\label{fig4}
\end{figure}

As shown in Figure \ref{fig:results_human2}, there are significant differences in the distributions between Detail Description data and Referring QA  data. The Detail Description data are widely distributed in the upper right corner of the space, while the Referring QA data are widely distributed in the lower left corner of the space. This indicates that the task type used as a prompt can significantly influence the attributes of the data and lead to differences in quality.

\begin{wrapfigure}{r}{0.4\textwidth}
    \vspace{-15pt}
    \includegraphics[width=\linewidth]{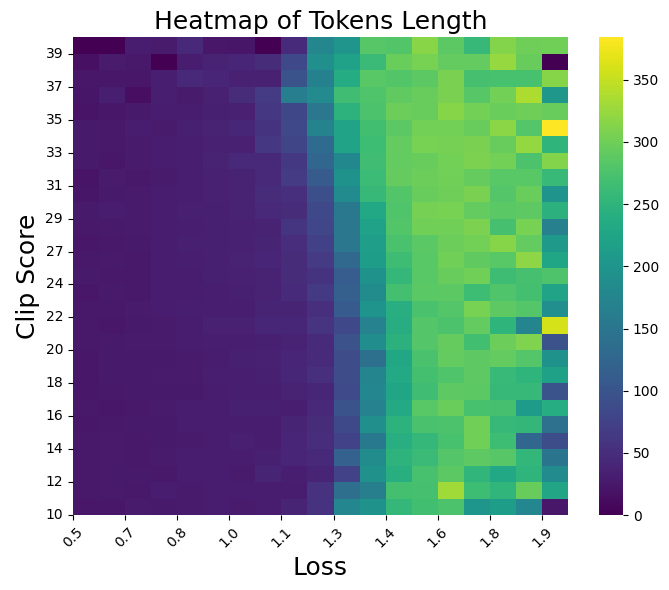}
    \caption{Heatmap visualization of statistics on token length.}
    \label{fig:analysis}
    \vspace{-15pt}
\end{wrapfigure}

Meanwhile, as shown in Figure \ref{fig:results_human3},  we found that the data quality distribution of Detail Description and Complex Reasoning tasks also differs significantly. In particular, data quality distribution for Complex Reasoning tasks, which are constructed through multi-turn dialogues, is spread over a wider area, highlighting the challenge of maintaining data quality in dialogue-based task types.



Additionally, various factors can influence the data's attributes and quality besides the task type. We analyzed the token lengths of data in all regions and visualized the distribution using a heatmap. As shown in Figure \ref{fig:analysis}, brighter areas indicate longer text lengths, primarily on the right side, suggesting a consistent correlation between token length and data loss. This indicates that visual instruction data created using LLMs from the same representation space have a stable logical hierarchy and rich information. Therefore, longer visual instruction data can effectively improve data quality by providing more detailed and coherent information.

\section{Conclusion}  

In this paper, we introduce a curriculum learning method that imitates the human learning process. By gradually improving the quality of training data from easy to difficult stages, our method enhances performance while requiring less training data. In addition, we demonstrate the effectiveness of utilizing a dual-attribute representation space in controlling the quality of multimodal training data that divides data subsets based on clip score and model loss. We find that not only do data with higher dual-attribute values lead to better performance, but we also found a correlation between the task type used during visual instruction data creation and the distribution of positions in the dual-attribute space. At the same time, we found that different selection strategies for the subset of high-quality data are needed at different stages of training. When MLLMs have completed instruction fine-tuning tasks, incorporating curriculum learning can significantly improve fine-tuning performance.

\paragraph{Limitation} Regarding the limitations, the scope of our experiments was constrained by computational costs, limiting our focus to a single model. We conducted all related experiments on LLaVA-v1.5\cite{llava1.5}. Nevertheless, this approach allowed us to achieve significant results within our computational constraints. To address these limitations, future research could explore more powerful and advanced models as computational resources allow.



\appendix

\bibliography{neurips_2024}
\bibliographystyle{plain}

\paragraph{Evaluation Metrics}

To better validate the activation of the model's conversational capabilities, the experiments were centered around the evaluation of open questions. The evaluation metrics mainly use accuracy for VQA tasks. We compare a model optimized through continuous instruction tuning with TA-selected-15k against leading MLLMs.: BLIP-2 \cite{Blip-2}, InstructBLIP \cite{instructblip}, Shikra \cite{shikra}, IDEFICS \cite{idefics}, Qwen-VL(-Chat) \cite{bai2023qwen}, mPLUG-Owl2 \cite{ye2023mplugowl2} and LLaVA-v1.5 \cite{llava1.5}.  We evaluate these models on popular benchmarks: VQA-v2 \cite{goyal2017making}, GQA \cite{GQA2019}, VisWiz \cite{gurari2018vizwiz}, ScienceQA-IMG \cite{lu2022learn}, TextVQA \cite{singh2019towards}. 


\newpage


\clearpage

\newpage
\section*{NeurIPS Paper Checklist}


\begin{enumerate}

\item {\bf Claims}
    \item[] Question: Do the main claims made in the abstract and introduction accurately reflect the paper's contributions and scope?
    \item[] Answer: \answerYes{}
    \item[] Justification: see Section 1.

\item {\bf Limitations}
    \item[] Question: Does the paper discuss the limitations of the work performed by the authors?
    \item[] Answer:  \answerYes{} 
    \item[] Justification: see the conclusion.

\item {\bf Theory Assumptions and Proofs}
    \item[] Question: For each theoretical result, does the paper provide the full set of assumptions and a complete (and correct) proof?
    \item[] Answer: \answerYes{}
    \item[] Justification: see Section 3.

    \item {\bf Experimental Result Reproducibility}
    \item[] Question: Does the paper fully disclose all the information needed to reproduce the main experimental results of the paper to the extent that it affects the main claims and/or conclusions of the paper (regardless of whether the code and data are provided or not)?
    \item[] Answer: \answerYes{}  
    \item[] Justification: see Section 4.

\item {\bf Open access to data and code}
    \item[] Question: Does the paper provide open access to the data and code, with sufficient instructions to faithfully reproduce the main experimental results, as described in supplemental material?
    \item[] Answer: \answerYes{} 
    \item[] Justification: see abstract.

\item {\bf Experimental Setting/Details}
    \item[] Question: Does the paper specify all the training and test details (e.g., data splits, hyperparameters, how they were chosen, type of optimizer, etc.) necessary to understand the results?
    \item[] Answer: \answerYes{} 
    \item[] Justification: see Section 4.

\item {\bf Experiment Statistical Significance}
    \item[] Question: Does the paper report error bars suitably and correctly defined or other appropriate information about the statistical significance of the experiments?
    \item[] Answer:  \answerNo{} 
    \item[] Justification: error bars are not reported because it would be too computationally expensive

\item {\bf Experiments Compute Resources}
    \item[] Question: For each experiment, does the paper provide sufficient information on the computer resources (type of compute workers, memory, time of execution) needed to reproduce the experiments?
    \item[] Answer: \answerYes{} 
    \item[] Justification: see Section 4.
    
\item {\bf Code Of Ethics}
    \item[] Question: Does the research conducted in the paper conform, in every respect, with the NeurIPS Code of Ethics \url{https://neurips.cc/public/EthicsGuidelines}?
    \item[] Answer: \answerYes{}. 
    \item[] Justification: see Section 4.

\item {\bf Broader Impacts}
    \item[] Question: Does the paper discuss both potential positive societal impacts and negative societal impacts of the work performed?
    \item[] Answer: \answerNA{} 
    \item[] Justification: there is no societal impact of the work performed.
    
\item {\bf Safeguards}
    \item[] Question: Does the paper describe safeguards that have been put in place for responsible release of data or models that have a high risk for misuse (e.g., pretrained language models, image generators, or scraped datasets)?
    \item[] Answer: \answerYes{}. 
    \item[] Justification: see Section 4.

\item {\bf Licenses for existing assets}
    \item[] Question: Are the creators or original owners of assets (e.g., code, data, models), used in the paper, properly credited and are the license and terms of use explicitly mentioned and properly respected?
    \item[] Answer: \answerYes{}. 
    \item[] Justification: see Section 4.

\item {\bf New Assets}
    \item[] Question: Are new assets introduced in the paper well documented and is the documentation provided alongside the assets?
    \item[] Answer: \answerNA{} 
    \item[] Justification:  the paper does not release new assets.

\item {\bf Crowdsourcing and Research with Human Subjects}
    \item[] Question: For crowdsourcing experiments and research with human subjects, does the paper include the full text of instructions given to participants and screenshots, if applicable, as well as details about compensation (if any)? 
    \item[] Answer: \answerNA{} 
    \item[] Justification: the paper does not involve crowdsourcing nor research with human subjects.

\item {\bf Institutional Review Board (IRB) Approvals or Equivalent for Research with Human Subjects}
    \item[] Question: Does the paper describe potential risks incurred by study participants, whether such risks were disclosed to the subjects, and whether Institutional Review Board (IRB) approvals (or an equivalent approval/review based on the requirements of your country or institution) were obtained?
    \item[] Answer: \answerNA{} 
    \item[] Justification: the paper does not involve crowdsourcing nor research with human subjects.

\end{enumerate}

\end{document}